\documentclass[11pt]{article}

\usepackage[final]{acl}

\usepackage{times}
\usepackage{latexsym}
\usepackage[T1]{fontenc}
\usepackage[utf8]{inputenc}
\usepackage{microtype}
\IfFileExists{inconsolata.sty}{\usepackage{inconsolata}}{}

\usepackage{graphicx}
\usepackage{booktabs}
\usepackage{enumitem}
\usepackage{array}
\usepackage{multirow}
\usepackage{amsmath}
\usepackage{mathtools}  
\usepackage{amssymb}
\usepackage{fontspec}
\usepackage{fontspec}

\newfontfamily\cjkfont{gbsn00lp.ttf}

\newcommand{\zh}[1]{{\cjkfont #1}}

\XeTeXlinebreaklocale "zh"
\usepackage{tikz}
\usetikzlibrary{positioning, fit, backgrounds, calc, arrows.meta}
\usepackage{colortbl}
\usepackage{url}

\renewcommand{\textfraction}{0.20}

\definecolor{intracol}{RGB}{15,109,100}  
\definecolor{sentcol}{RGB}{91,63,160}    
\definecolor{monocol}{RGB}{154,165,173}  
\definecolor{langEN}{RGB}{42,111,214}    
\definecolor{langNL}{RGB}{232,133,12}    
\definecolor{langZH}{RGB}{209,52,47}     
\newcommand{\embEN}[1]{\textcolor{langEN}{#1}}
\newcommand{\embNL}[1]{\textcolor{langNL}{#1}}
\newcommand{\embZH}[1]{\textcolor{langZH}{#1}}
\newcommand{\dtag}[1]{{\scriptsize\textsf{\textcolor{black!55}{#1}}}}  

\tikzset{
  wordfill/.style={fill=intracol!80},
  sentfill/.style={fill=sentcol!80},
  monofill/.style={fill=monocol!35},
  langchip/.style 2 args={draw=#1!70, fill=#1!12, text=#1, rounded corners=2.5pt,
    inner xsep=5pt, inner ysep=2.5pt, text height=1.55ex, text depth=.35ex,
    minimum width=2.3em, align=center, font=\footnotesize\bfseries},
}
\newcommand{\panelhdr}[2]{%
  \begin{tikzpicture}[x=1cm,y=1cm]
    \node[fill=black!8, draw=black!25, line width=0.4pt, text=black!82,
          rounded corners=2.5pt, inner xsep=5pt, inner ysep=2pt,
          font=\footnotesize\bfseries, anchor=west] (b) at (0,0) {#1};
    \node[anchor=west, font=\footnotesize\bfseries, text=black!70]
          at ($(b.east)+(0.16,0)$) {#2};
  \end{tikzpicture}%
}
\newsavebox{\figonelegend}   

\title{Feeding BabyLMs Macaroni: Code-Switching Curricula Cause Cross-Lingual Convergence}

\author{
  \textbf{Dries Rooryck}\textsuperscript{$*$1} \quad
  \textbf{Alex Cai}\textsuperscript{$*$1} \quad
  \textbf{Yonatan Belinkov}\textsuperscript{1,2} \quad
  \textbf{David Alvarez-Melis}\textsuperscript{1} \quad
  \textbf{Kiant\'e Brantley}\textsuperscript{1} \\
  [4pt]
  \textsuperscript{1}Kempner Institute, Harvard University, Cambridge, MA, USA \\
  \textsuperscript{2}Technion -- Israel Institute of Technology, Haifa, Israel \\[4pt]
  {\small\texttt{dries\_rooryck@college.harvard.edu, adzcai@g.harvard.edu}}
}

\begin{document}
\raggedbottom
\maketitle
\begingroup\renewcommand\thefootnote{*}\footnotetext{Equal contribution.}\endgroup

\begin{abstract}
Children in multilingual communities often \emph{code-switch}, using multiple languages in a single utterance. Can we induce cross-lingual alignment in language models by training on code-switched text? We pretrain small decoder-only transformers on two $100$M-word multilingual corpora: a base corpus formed by mixing the English, Dutch, and Chinese BabyBabelLM datasets, and a corpus generated from it by inserting word- and sentence-level code-switching using an LLM. We find that training on code-switched data aligns the representations of parallel text, particularly across different scripts, and that this alignment persists through training on monolingual documents. Under a learning curriculum that progresses from word-level code-switching, to sentence-level code-switching, to monolingual documents, models trained on code-switched data outperform baselines trained without it on the BabyLM evaluation suite. Our work characterizes code-switching curriculum learning as an effective data augmentation method for multilingual pretraining. We release our code, data, and models at \url{https://github.com/drooryck/multilingual-macaroni}.
\end{abstract}

\section{Introduction}
\label{sec:intro}

Modern large language models (LLMs) are trained on orders of magnitude more language data than a person typically is exposed to in their lifetime.
Yet children in any country communicate fluently, often in multiple languages,
after hearing under the equivalent of just 100 million English words \citep{choshen2026babylm}.
The BabyLM competition seeks to train computational language models under such developmentally plausible data constraints.
This year's multilingual track asks participants to prepare the highest possible performing model across English, Dutch, and Chinese using the above total word budget.

\begin{figure}[t!]
\centering
\panelhdr{A}{Examples of each data type}

\vspace{1pt}
\tikzset{
  panelhead/.style={fill=#1, rounded corners=2pt, minimum width=7.3cm,
    inner ysep=2.5pt, font=\scriptsize\bfseries, text=black!75},
  exlab/.style={font=\tiny, inner sep=1.2pt, fill=white, text=black!80},
  csarrow/.style={-{Stealth[length=4pt,width=3.5pt]}, line width=0.7pt},
}

\begin{tikzpicture}[x=1cm,y=1cm]
  \node[panelhead=monocol!30] at (3.7,0.75) {Non-CS data};
  \node[langchip={langEN}{}] (en) at (1.2,0) {en};
  \node[langchip={langNL}{}] (nl) at (3.7,0) {nl};
  \node[langchip={langZH}{}] (zh) at (6.2,0) {zh};
  \node[font=\tiny, text=langEN, align=center] at (1.2,-0.55) {That's illegal!};
  \node[font=\tiny, text=langNL, align=center] at (3.7,-0.55) {Geef me een kus .};
  \node[font=\tiny, text=langZH, align=center] at (6.2,-0.55) {\zh{有什么区别}?};
\end{tikzpicture}

\vspace{1pt}

\begin{tikzpicture}[x=1cm,y=1cm]
  \node[panelhead=sentcol!22] at (3.7,1.0) {Sentence-level CS \,---\, 6 ordered directions};
  \coordinate (EN) at (0.5,0);
  \coordinate (ZH) at (6.9,0);
  \coordinate (NL) at (3.7,-1.7);
  \begin{scope}[transform canvas={shift={(0,0.16)}}]
    \draw[csarrow, langZH] ($(ZH)!0.10!(EN)$) -- node[exlab]
      {\textcolor{langEN}{It's you.} \textcolor{langZH}{\zh{不，不是。}}} ($(ZH)!0.90!(EN)$);
  \end{scope}
  \begin{scope}[transform canvas={shift={(0,-0.16)}}]
    \draw[csarrow, langEN] ($(EN)!0.10!(ZH)$) -- node[exlab]
      {\textcolor{langZH}{\zh{他是谁呀。}} \textcolor{langEN}{It's Xu Song.}} ($(EN)!0.90!(ZH)$);
  \end{scope}
  \begin{scope}[transform canvas={shift={(-0.156,-0.111)}}]
    \draw[csarrow, langEN] ($(EN)!0.12!(NL)$) -- node[exlab, sloped]
      {\textcolor{langNL}{Ken je hem?} \textcolor{langEN}{He's my brother.}} ($(EN)!0.88!(NL)$);
  \end{scope}
  \begin{scope}[transform canvas={shift={(0.156,0.111)}}]
    \draw[csarrow, langNL] ($(NL)!0.12!(EN)$) -- node[exlab, sloped]
      {\textcolor{langEN}{Where's the dog?} \textcolor{langNL}{Hij slaapt.}} ($(NL)!0.88!(EN)$);
  \end{scope}
  \begin{scope}[transform canvas={shift={(0.156,-0.111)}}]
    \draw[csarrow, langZH] ($(ZH)!0.12!(NL)$) -- node[exlab, sloped]
      {\textcolor{langNL}{Zie je hem ?} \textcolor{langZH}{\zh{他在那里。}}} ($(ZH)!0.88!(NL)$);
  \end{scope}
  \begin{scope}[transform canvas={shift={(-0.156,0.111)}}]
    \draw[csarrow, langNL] ($(NL)!0.12!(ZH)$) -- node[exlab, sloped]
      {\textcolor{langZH}{\zh{他妈姓杨.}} \textcolor{langNL}{Ik heet Li.}} ($(NL)!0.88!(ZH)$);
  \end{scope}
  \node[langchip={langEN}{}] at (EN) {en};
  \node[langchip={langZH}{}] at (ZH) {zh};
  \node[langchip={langNL}{}] at (NL) {nl};
\end{tikzpicture}

\vspace{5pt}

\begin{tikzpicture}[x=1cm,y=1cm]
  \node[panelhead=intracol!42] at (3.7,1.0) {Word-level CS \,---\, 6 ordered directions};
  \coordinate (EN) at (0.5,0);
  \coordinate (ZH) at (6.9,0);
  \coordinate (NL) at (3.7,-1.7);
  \begin{scope}[transform canvas={shift={(0,0.16)}}]
    \draw[csarrow, langZH] ($(ZH)!0.10!(EN)$) -- node[exlab]
      {I know it'll \textcolor{langZH}{\zh{行}}.} ($(ZH)!0.90!(EN)$);
  \end{scope}
  \begin{scope}[transform canvas={shift={(0,-0.16)}}]
    \draw[csarrow, langEN] ($(EN)!0.10!(ZH)$) -- node[exlab]
      {\zh{得}\,\textcolor{langEN}{believe}\,\zh{爱情。}} ($(EN)!0.90!(ZH)$);
  \end{scope}
  \begin{scope}[transform canvas={shift={(-0.156,-0.111)}}]
    \draw[csarrow, langEN] ($(EN)!0.12!(NL)$) -- node[exlab, sloped]
      {Pak die \textcolor{langEN}{shovel} .} ($(EN)!0.88!(NL)$);
  \end{scope}
  \begin{scope}[transform canvas={shift={(0.156,0.111)}}]
    \draw[csarrow, langNL] ($(NL)!0.12!(EN)$) -- node[exlab, sloped]
      {That's \textcolor{langNL}{illegaal}!} ($(NL)!0.88!(EN)$);
  \end{scope}
  \begin{scope}[transform canvas={shift={(0.156,-0.111)}}]
    \draw[csarrow, langZH] ($(ZH)!0.12!(NL)$) -- node[exlab, sloped]
      {Geef me een \textcolor{langZH}{\zh{吻}} .} ($(ZH)!0.88!(NL)$);
  \end{scope}
  \begin{scope}[transform canvas={shift={(-0.156,0.111)}}]
    \draw[csarrow, langNL] ($(NL)!0.12!(ZH)$) -- node[exlab, sloped]
      {\zh{有什么}\,\textcolor{langNL}{verschil}?} ($(NL)!0.88!(ZH)$);
  \end{scope}
  \node[langchip={langEN}{}] at (EN) {en};
  \node[langchip={langZH}{}] at (ZH) {zh};
  \node[langchip={langNL}{}] at (NL) {nl};
\end{tikzpicture}

\vspace{2pt}
\panelhdr{B}{Corpus composition}

\vspace{1pt}
\tikzset{seg/.style={draw=black!45, line width=0.4pt, rounded corners=1.8pt},
         divider/.style={draw=black!40, line width=0.4pt},
         ctitle/.style={font=\scriptsize\bfseries, text=black!75},
         leg/.style={font=\tiny, text=black!80, anchor=west, inner sep=1pt}}
\sbox{\figonelegend}{%
  \begin{tikzpicture}[x=1cm,y=1cm]
    \draw[seg, wordfill] (0,-0.11) rectangle (0.24,0.11);
    \node[leg] (l1) at (0.34,0) {word-level CS};
    \draw[seg, sentfill] ($(l1.east)+(0.36,-0.11)$) rectangle ($(l1.east)+(0.60,0.11)$);
    \node[leg] (l2) at ($(l1.east)+(0.70,0)$) {sentence-level CS};
    \draw[seg, monofill] ($(l2.east)+(0.36,-0.11)$) rectangle ($(l2.east)+(0.60,0.11)$);
    \node[leg] (l3) at ($(l2.east)+(0.70,0)$) {non-CS};
  \end{tikzpicture}%
}
\begin{tikzpicture}[x=1cm,y=1cm]
  \def\bh{0.30}   
  \def\rs{0.46}   
  \def\uw{1.95}   
  \def\cx{2.65}   
  \def\cw{1.95}   
  \node[langchip={langEN}{}] at (-0.45,{2*\rs+\bh/2}) {en};
  \node[langchip={langNL}{}] at (-0.45,{1*\rs+\bh/2}) {nl};
  \node[langchip={langZH}{}] at (-0.45,{0*\rs+\bh/2}) {zh};
  \foreach \i in {0,1,2} { \draw[seg, monofill] (0,\i*\rs) rectangle (\uw,\i*\rs+\bh); }
  \node[ctitle] at (\uw/2,{2*\rs+\bh+0.30}) {Non-CS corpus};
  \foreach \i in {0,1,2} {
    \begin{scope}
      \clip[rounded corners=1.8pt] (\cx,\i*\rs) rectangle (\cx+\cw,\i*\rs+\bh);
      \fill[wordfill] (\cx,\i*\rs)            rectangle (\cx+\cw/3,\i*\rs+\bh);
      \fill[sentfill] (\cx+\cw/3,\i*\rs)      rectangle (\cx+2*\cw/3,\i*\rs+\bh);
      \fill[monofill] (\cx+2*\cw/3,\i*\rs)    rectangle (\cx+\cw,\i*\rs+\bh);
    \end{scope}
    \draw[divider] (\cx+\cw/3,\i*\rs)   -- (\cx+\cw/3,\i*\rs+\bh);
    \draw[divider] (\cx+2*\cw/3,\i*\rs) -- (\cx+2*\cw/3,\i*\rs+\bh);
    \draw[seg, fill=none] (\cx,\i*\rs) rectangle (\cx+\cw,\i*\rs+\bh);
  }
  \node[ctitle] at (\cx+\cw/2,{2*\rs+\bh+0.30}) {CS corpus};
  \draw[-{Stealth[length=3pt,width=2.5pt]}, black!55, line width=0.4pt] (\cx,-0.17) -- (\cx+\cw,-0.17)
    node[midway, below=-0.5pt, font=\tiny, text=black!70] {curriculum training order};
  \node[anchor=south, inner sep=0pt] at ($(current bounding box.north)+(0,0.10)$) {\usebox{\figonelegend}};
\end{tikzpicture}
\caption{\textbf{Composition of our synthetic code-switched corpus.}
\textbf{A}: We synthesize code-switched (CS) data by prompting an LLM to translate words or sentences from documents in the monolingual competition datasets.
Examples shown are from our data.
\textbf{B}: We train models on two corpora.
Both are one-third English, Dutch, and Chinese.
The non-CS corpus comprises monolingual source documents,
while the CS corpus splits each language's subset equally into word-level CS, sentence-level CS, and monolingual documents,
shown left to right in curriculum training order.}
\label{fig:overview}
\end{figure}
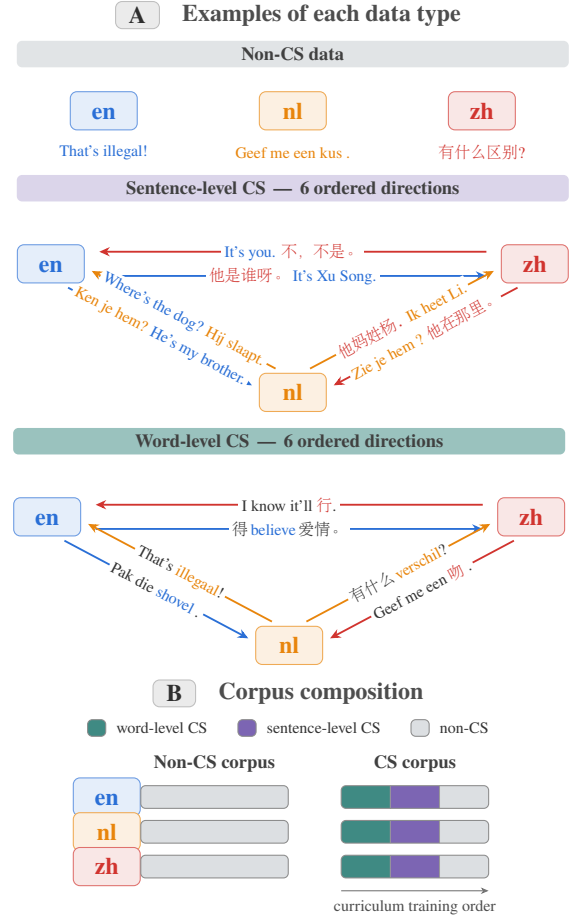

\begin{table}[t]
  \centering \small
  \begin{tabular}{@{}llrrrr@{}}
    \toprule
    matrix & emb. & Word & Sent. & Non-CS & Total \\
    \midrule
    \multirow{2}{*}{English} & $\to$NL & 5.2 & 6.0 & \multirow{2}{*}{11.1} & \multirow{2}{*}{33.4} \\
                             & $\to$ZH & 5.9 & 5.1 & & \\
    \addlinespace[2pt]
    \multirow{2}{*}{Dutch}   & $\to$EN & 5.8 & 6.3 & \multirow{2}{*}{11.7} & \multirow{2}{*}{35.1} \\
                             & $\to$ZH & 5.9 & 5.4 & & \\
    \addlinespace[2pt]
    \multirow{2}{*}{Chinese} & $\to$EN & 5.7 & 5.8 & \multirow{2}{*}{10.4} & \multirow{2}{*}{31.2} \\
                             & $\to$NL & 4.7 & 4.6 & & \\
    \midrule
    Total & & 33.2 & 33.2 & 33.2 & 99.6 \\
    \bottomrule
  \end{tabular}
  \caption{Composition of the CS corpus by \emph{matrix language} (the source document language) and \emph{embedded language} (the language of the inserted words or translated sentences)
  in millions of byte-premium-adjusted words (see Appendix~\ref{app:data}).
  Cells may not sum exactly to totals due to rounding.}
  \label{tab:composition}
\end{table}

We take inspiration from infants raised in multilingual environments, who,
within a single utterance, will use words from multiple languages,
a phenomenon known as \textbf{code-switching} (CS)
\citep{volterra1978acquisition}, or, rarely, ``macaronic language''.\footnote{Some authors use the terms ``code-switching'' and ``code-mixing'' for different phenomena.
We follow \citet{yoo2025cscl} in distinguishing between word-level and sentence-level code-switching.
See also \url{https://en.wikipedia.org/wiki/Macaronic\_language}.}
While children code-switch naturally during language acquisition,
might training neural language models intentionally on CS data help them align their representations of parallel text?
This hypothesis has driven numerous schemes for multilingual pretraining \citep{yang2020alm, li2024prealign, wang2025scaling} and fine-tuning \citep{qin2020cosdaml, zheng2024lexicon, yoo2025cscl, wang2026reasoning}.

Training on CS data, however, introduces a complication:
the trained model is not typically intended to generate CS text, but rather monolingual text in each of multiple languages.
Code-switching curriculum learning \citep{yoo2025cscl} presents a promising solution:
first train on word-level CS, then sentence-level CS, and finally monolingual documents.
This ordering imitates the stages of a human learning a new language:
first one learns new words, then incorporates whole sentences into their speech,
and finally speaks fully in the new language.

Our work adapts code-switching curriculum learning to data-constrained multilingual pretraining.
We investigate the causal effects of pretraining on CS data
by comparing against baselines trained on the same documents kept entirely in the original language.
Our research is organized along the following questions:
\begin{enumerate}[noitemsep, topsep=3pt, leftmargin=*]
\item Does CS training improve downstream performance on the BabyLM evaluation suite?
\item Does CS training align the model's internal representations across languages?
\item Does the model represent words seen embedded in CS contexts differently than words seen only in monolingual contexts?
\end{enumerate}
Our contributions, respectively, are that for models trained on CS data:
\begin{enumerate}[noitemsep, topsep=3pt, leftmargin=*]
\item Performance on the BabyLM evaluation suite improves when trained under a curriculum but not under shuffled ordering (\S\ref{sec:effect});
\item Representations of parallel text align,
and under curriculum ordering,
the alignment persists through training on monolingual documents (\S\ref{sec:mechanism});
\item Embedded words align more closely to their translations than words that only appear in monolingual contexts (\S\ref{sec:words}).
\end{enumerate}
We release our models\footnote{\url{https://huggingface.co/drooryck/multilingual-macaroni-models}},
corpora\footnote{\url{https://huggingface.co/datasets/drooryck/multilingual-macaroni-corpus}},
and code\footnote{\url{https://github.com/drooryck/multilingual-macaroni}}.

\section{Related Work}
\label{sec:related}

\paragraph{Developmentally plausible language modeling.}
The first three iterations of the BabyLM challenge \citep{warstadt2023babylm, hu2024babylm2, charpentier2025babylm3} established data-constrained language acquisition as an interdisciplinary question of interest to machine learning researchers, linguists, and cognitive scientists.
Relevant past submissions include a negative finding for cognitively-inspired curriculum learning \citep{diehlmartinez2023climb} and a benchmark for detection of ungrammaticality, including unnatural CS, in second-language acquisition \citep{gao2025bliss}.
\citet{zeng2026bilingual} also study the effect of various synthetic data mixtures on bilingual BabyLMs.
Our work adopts the motivation, constraints, and evaluation suite of this year's challenge \citep{choshen2026babylm},
which introduces a multilingual track based on the BabyBabelLM corpora \citep{jumelet2026babybabellm}.

\paragraph{CS as data augmentation for multilingual training.}
While attempts to model human CS predate the deep learning era \citep{chan2006codemixing, franco2007spanglish},
recent works generate CS as a data augmentation strategy for multilingual training.
Some works procedurally generate CS data using bilingual dictionaries \citep{qin2020cosdaml, zhu2023sogo, zheng2024lexicon, feng2022limitation}.
Others prompt LLMs, as we do,
which exhibit more naturalistic patterns \citep{kuwanto2026equivalence, wang2025scaling}.
While the above works demonstrate benefits of CS training,
\citet{shao2026mixed} suggest that CS data is less important than parallel data for a model's ability to translate
(but that, surprisingly, neither kind of multilingual data is required for other cross-lingual understanding and reasoning tasks).
However, their category of CS documents does not match our definitions of word- and sentence-level CS,
and ultimately both our works stress the importance of fine-grained lexical alignment for translation.
Other work has applied CS for cross-lingual transfer at inference time \citep{yoo2026gradual}, in chain of thought \citep{wang2026reasoning}, or for instruction tuning \citep{asano2026beyond}.
Our work most closely follows code-switching curriculum learning \citep{yoo2025cscl}.
Whereas code-switching curriculum learning seeks to fine-tune a pretrained model,
our study focuses on pretraining in the data-constrained regime and attempts a closer analysis of model representations.

\paragraph{Cross-lingual representation alignment.}
Recent papers characterize LLMs' ability to generalize across languages despite being trained on mostly monolingual documents.
Interpretability studies find shared model components or subspaces across languages \citep{conneau2020emerging, dufter2020identifying, chang2022geometry}
or investigate how cross-lingual alignment emerges throughout the training process \citep{blevins2022analyzing, wang2024probing}.
Token overlap between languages tends to improve cross-lingual generalization \citep{pires2019multilingual, kallini2025falsefriends},
as does typological similarity \citep{muller2023languages,longpre2026atlas}.
See \citet{hammerl2024survey} for a survey on cross-lingual alignment.
Our work contributes a controlled study of how code-switching curriculum learning affects cross-lingual alignment.

\section{Synthetic Code-Switched Data Generation}
\label{sec:data}

\begin{table*}[t!]
  \centering \small
  \setlength{\tabcolsep}{4pt}
  \resizebox{\textwidth}{!}{%
  \begin{tabular}{@{}ll c cc c ccccc@{}}
    \toprule
    & & & & & & \multicolumn{5}{c}{Selected tasks (accuracy, avg.\ over languages)} \\
    \cmidrule(lr){7-11}
    Ordering & Corpus & Leaderboard & Fine-tune avg. & Zero-shot avg. & & SIB-200 & INCLUDE & Global PIQA & BMLAMA & MultiBLiMP \\
    \midrule
    \multirow{2}{*}{Shuffled}& non-CS     & 46.44$\pm$0.25 & 42.27$\pm$0.48 & 51.77$\pm$0.17 & & 71.85$\pm$1.64 & 29.02$\pm$2.61 & 36.84$\pm$1.24 & 21.60$\pm$2.56 & 86.91$\pm$0.72 \\
                            & CS        & 46.55$\pm$0.47 & 42.33$\pm$0.79 & 51.95$\pm$0.14 & & \textbf{73.33$\pm$1.23} & 28.96$\pm$2.13 & 35.86$\pm$0.69 & 19.87$\pm$1.97 & \textbf{88.05$\pm$0.87} \\
    \midrule
    \multirow{2}{*}{Curriculum}& non-CS     & 46.36$\pm$0.37 & 42.26$\pm$0.74 & 51.60$\pm$0.22 & & 72.33$\pm$1.83 & 27.79$\pm$3.33 & 36.85$\pm$0.60 & 21.11$\pm$1.56 & 86.52$\pm$0.46 \\
                            & CS        & \textbf{46.72$\pm$0.25} & \textbf{42.81$\pm$0.42} & 51.72$\pm$0.13 & & \textbf{74.27$\pm$1.20} & 29.13$\pm$1.56 & 36.66$\pm$0.78 & 20.31$\pm$2.42 & \textbf{87.17$\pm$0.50} \\
    \midrule
    \multirow{4}{*}{\shortstack[l]{Shuffled\\(controls)}}
                                & word-level only       & 46.59$\pm$0.24 & 42.60$\pm$0.38 & 51.68$\pm$0.25 & & 73.44$\pm$1.70 & 28.63$\pm$2.54 & 35.79$\pm$1.08 & 20.96$\pm$1.80 & 87.02$\pm$0.73 \\
                                & sentence-level only   & 46.70$\pm$0.29 & 42.60$\pm$0.45 & 51.96$\pm$0.25 & & 72.52$\pm$1.48 & 29.19$\pm$1.91 & 36.25$\pm$1.53 & 21.35$\pm$1.35 & 87.92$\pm$0.64 \\
                                & document translation  & 46.56$\pm$0.38 & 42.43$\pm$0.63 & 51.83$\pm$0.24 & & 72.75$\pm$0.94 & 28.57$\pm$3.60 & 36.20$\pm$0.79 & 20.02$\pm$1.97 & 87.89$\pm$0.75 \\
                                & word salad            & 46.08$\pm$0.34 & 42.14$\pm$0.63 & 51.07$\pm$0.20 & & 72.42$\pm$1.90 & 28.13$\pm$2.24 & 36.39$\pm$0.78 & 18.47$\pm$2.69 & 85.82$\pm$0.87 \\
    \bottomrule
  \end{tabular}}
  \caption{Accuracy on the BabyLM evaluation suite (mean\,$\pm$\,SD) across eight seeds per condition.
  We select five tasks:
  the two language-averaged tasks with the largest positive gap between the curriculum CS and non-CS models (SIB-200 $+1.94$, INCLUDE $+1.34$),
  the two with the most negative (BMLAMA $-0.80$, Global PIQA $-0.20$),
  and MultiBLiMP as a grammatical reference task.
   Bold cells mark statistical significance on a two-sided paired $t$-test between the CS and non-CS models (uncorrected for multiple hypotheses).
  Full per-task results are in Table~\ref{tab:sweep} (Appendix~\ref{app:evalsuite}).}
  \label{tab:results}
\end{table*}

We construct two distinct corpora that both satisfy BabyLM's data budget: one with code-switched (CS) data and the other without (non-CS).
Figure~\ref{fig:overview} illustrates their composition
and Figure~\ref{fig:composition} (Appendix~\ref{app:data}) gives longer examples.
See Appendix~\ref{app:data} for detailed accounting.
We also create four additional corpora as experimental controls.

\paragraph{Monolingual-document (non-CS) corpus.}
We begin with the English, Dutch, and Chinese BabyBabelLM datasets \citep{jumelet2026babybabellm}.
We sample a third of each and combine them to obtain our base non-CS corpus of monolingual documents.

\paragraph{CS corpus.}
We begin with the English third of the non-CS corpus (33.4M words) and separate it further into thirds:
one kept monolingual, one for word-level CS (where English is the matrix language), and one for sentence-level CS (where half of the sentences in each document are translated).
The monolingual third is kept as-is. The word-level third is divided into one half for Dutch insertions and one half for Chinese insertions, where we use an instruction-tuned LLM to translate a subset of words to the embedded language.
We similarly bisect the sentence-level third
and instruct the LLM to translate, in-place, roughly half of the sentences in each document to the embedded language.
We then repeat this \emph{mutatis mutandis}  for the Dutch and Chinese subsets. See Table~\ref{tab:composition} for an exact breakdown of the final corpus composition.
We verify the quality of generated code-switched text through programmatic checks and through manual verification of sample documents by bilingual English-Dutch and English-Chinese speakers.
See Appendix~\ref{app:data} for our quality verification pipeline and Appendix~\ref{app:prompts} for LLM choice and prompts.

\paragraph{Corpora for controls.}
We construct four corpora to isolate the contribution of various experimental factors.
\begin{enumerate}
  \item \textbf{Word-level CS only}: we revert the sentence-level CS third of the CS corpus to the original monolingual documents.
  \item \textbf{Sentence-level CS only}: we similarly revert the word-level CS third.
  \item \textbf{Word salad}: for a third of the original non-CS corpus, we replace roughly $35\%$ of tokens of three or more letters, selected randomly, with words sampled from the embedded language's monolingual data. For Chinese source text, we insert words instead of replacing.
  \item \textbf{Document translation}: we take a third of documents in the original non-CS corpus and append to them an LLM-generated translation. We then discard a third of these documents to keep the word budget consistent.
\end{enumerate}
See Appendix~\ref{app:data} for verbatim examples from the word salad and document translation controls.

\section{Experimental Setup}
\label{sec:setup}

\begin{figure*}[t]
\centering
\includegraphics[width=\textwidth]{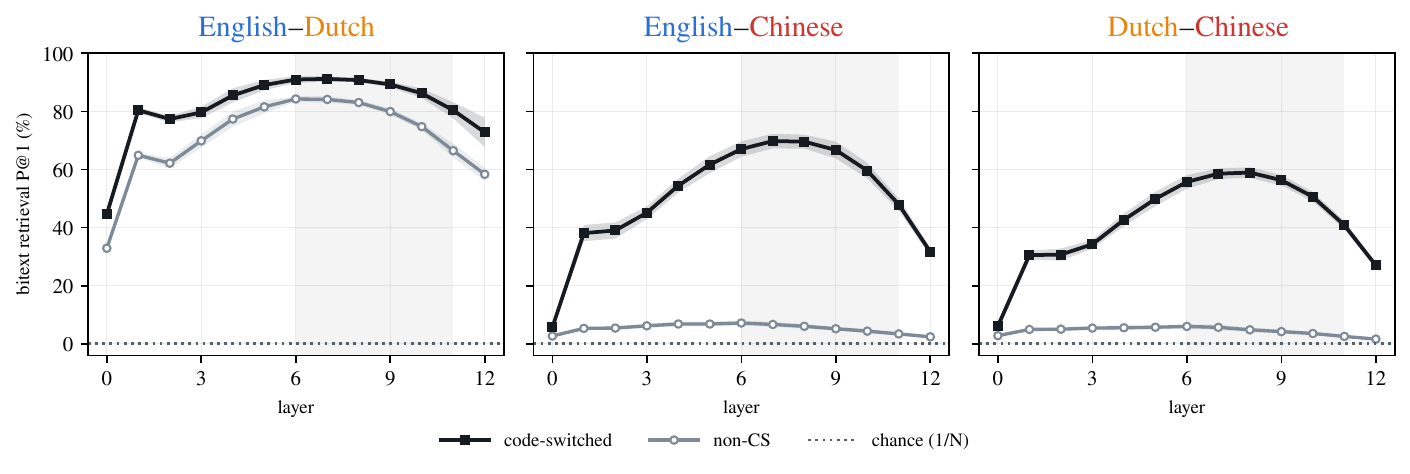}
\caption{Layerwise bitext retrieval precision@1 (\S\ref{sec:mechanism}) on $997$ parallel sentences from FLORES+ across English, Dutch, and Chinese.
We compare CS models against non-CS models (ignoring data ordering) and plot the mean $\pm1$~SD across the $16$ runs per group.
Retrieval in non-CS models fails for language pairs that differ in script.
See Figure~\ref{fig:retrieval-medrank} (Appendix~\ref{app:retrieval}) for the median percentile rank of the translation out of the candidate set.
Shading marks layers $6$--$11$, which the reported P@1 averages in other experiments are taken over.}
\label{fig:retrieval}
\end{figure*}

Our main experiments vary the training corpus (CS vs non-CS) and the data ordering (shuffled vs a three-stage curriculum).
Including our four control corpora yields $8$ distinct training setups in total.
For each setup, we train $8$ random seeds,
which affect the model initialization and document order.

\paragraph{Architecture and tokenizer.}
All models use the GPT-2-small architecture from the BabyLM multilingual baseline\footnote{\url{https://github.com/babylm-org/multilingual-training}}
(12 layers, hidden size 768, 12 heads, context length 1024).
We train a byte-level BPE tokenizer of vocabulary size $16,384$, matching the baseline implementation, on the monolingual-document corpus.

\paragraph{No padding.}
The BabyBabelLM corpora contain many short documents,
so to avoid excessive padding tokens,
we preprocess by concatenating all documents (separated by end-of-text tokens)
and splitting according to the model's context length.

\paragraph{Optimizer.}
Following the competition baselines,
we train models using Adam \citep{kingma2015adam} without weight decay,
a batch size of $16$,
and a learning rate schedule that starts at $0$,
warms up for $1\%$ of total training steps to $5\times10^{-5}$,
and cosine-decays to zero.

\paragraph{Curriculum learning.}
When training on the CS corpus, we compare between a learning curriculum and the default ordering ($10$ epochs on the shuffled CS corpus).
The curriculum consists of training first for 10 epochs over the word-level CS third of our CS corpus,
then 10 epochs on the sentence-level CS third,
and finally 10 epochs on the non-CS third,
resetting the learning rate schedule in each stage.
As for the baselines trained on the non-CS corpus,
we also apply these two orderings and their respective learning rate schedules,
maintaining the document IDs but using the original monolingual documents instead of the synthesized CS versions.
This results in four (corpus, ordering) pairs:
CS/shuffled, CS/curriculum, non-CS/shuffled, non-CS/curriculum.
The models trained on our control corpora all follow shuffled ordering.

\paragraph{Evaluation.}

\begin{figure*}[t]
\centering
\includegraphics[width=\textwidth]{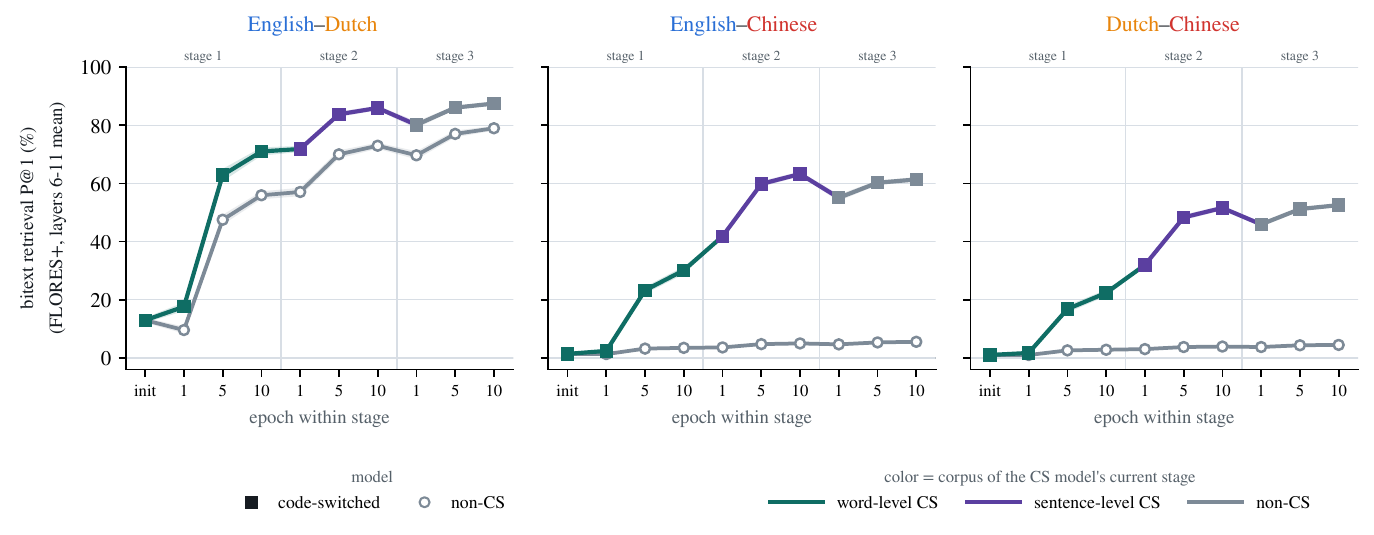}
\caption{Sentence-level alignment across the learning curriculum. Alignment improves throughout CS training and persists through training on monolingual documents.
The color of the CS model's line indicates the CS type of the current stage.}
\label{fig:retrievaltraj}
\end{figure*}

We evaluate models on the BabyLM evaluation suite,
which tests for grammatical fluency, cognitive plausibility, and model adaptability (via fine-tuning).
The full list of tasks is in Appendix~\ref{app:evalsuite}.
The English/Dutch/Chinese task suites differ slightly,
so all reported averages are of the per-language scores.

\section{Results}
\label{sec:effect}

Table~\ref{tab:results} summarizes performance on a subset of evaluation tasks.
We report per-task performance across the entire suite in Table~\ref{tab:sweep} (Appendix~\ref{app:evalsuite}) and also held-out bits per byte in Table~\ref{tab:heldout} (Appendix~\ref{app:evalsuite}).

\paragraph{Code-switching under a curriculum scores highest.}
Between our four corpus/ordering pairings,
we see that CS/curriculum performs highest on average (46.72),
followed by CS/shuffled, then non-CS/shuffled, and finally non-CS/curriculum.
Each of our training setups outperforms, on average across seeds, the BabyLM baseline \citep{choshen2026babylm},
which scores $45.94$ on the leaderboard\footnote{Model \texttt{BabyLM-2026-Baseline-GPT2-en\_nld\_zho\_equal} on the leaderboard at \url{https://huggingface.co/spaces/BabyLM-community/BabyLM-Leaderboard-2026}. Accessed 17 September 2026.}
We check for statistical significance with two-sided paired $t$-tests over all ${4 \choose 2} = 6$ pairwise comparisons of the ordering/corpus pairs,
applying Holm correction \citep{holm1979simple} for the familywise error rate.
Only the CS/curriculum improvement over non-CS/curriculum is significant ($p=0.016$);
the gap between CS/curriculum and non-CS/shuffled is not ($p=0.41$),
nor is any other comparison.

\paragraph{Fine-tuning tasks under the curriculum.}
For the models trained with a curriculum,
most of the CS models' advantage over the non-CS models comes from the fine-tuning tasks.
This suggests that the internal representations of CS models may be more generalizable toward various downstream tasks than those of models trained without CS.
We investigate further in \S\ref{sec:mechanism}.
However, this boost does not seem to hold under shuffled ordering,
where the CS and non-CS models' scores are within noise of each other for both zero-shot and fine-tuning averages.

\section{Cross-Lingual Representation Alignment}
\label{sec:mechanism}

Does training on code-switched data cause a model to align the representations of parallel sentences? We measure this by seeing whether a sentence and its translation have closer representations in CS models than in non-CS models.

\paragraph{Cross-lingual alignment metric.}
Formally, we measure a model's cross-lingual representation alignment using \textbf{bitext retrieval precision@1} (P@1)
\citep{artetxe2019massively,pires2019multilingual,hu2020xtreme}.
Given a sentence $s^A$ in language $A$,
and a list of sentences $s^B_1, \dots, s^B_N$ in language $B$ containing the translation of $s^A$,
we say the model successfully retrieves the translation if it is the original sentence's nearest neighbor in terms of the cosine similarity between their representations.
We represent a sentence by mean-pooling the residual stream vectors of its tokens.
We additionally apply cross-domain similarity local scaling \citep[CSLS,][]{lample2018word} to the cosine similarity values to account for the ``hubness'' issue of naive nearest neighbors.
Given a set of sentence pairs $(s^A_1, s^B_1), \dots, (s^A_N, s^B_N)$,
the bitext retrieval precision is the average success rate across all sentences in both directions.
We use $N{=}997$ parallel sentences across English, Dutch, and Chinese from the FLORES+ \texttt{dev} split \citep{nllb2024scaling, goyal2022flores}.
See Appendix~\ref{app:retrieval} for details.

\paragraph{Code-switched training aligns parallel sentences.}
Across all layers and language pairs, the CS model better aligns the sentence representations to their translations' representations (Figure~\ref{fig:retrieval}):
between English and Chinese under curriculum ordering,
the CS model retrieves the correct translation $61.4\%$ of the time (layers $6$--$11$, $8$-seed mean),
compared to only $5.5\%$ of the time in the non-CS baseline (Table~\ref{tab:controls-align}, Appendix~\ref{app:retrieval}).
A similar gap holds for bitext retrieval between Dutch and Chinese.
Other training-free measures,
including linear centered kernel alignment (CKA),
a sentence-level cosine gap,
centroid distance,
and the MEXA alignment score all corroborate this finding for the cross-script language pairs (Table~\ref{tab:alignment}, Appendix~\ref{app:retrieval}).

\paragraph{Coherence of CS matters.}
Must code-switching be coherent in order to improve alignment,
or is the gain driven simply by co-occurrence across tokens?
Our ``word salad'' control, where words are incoherently switched for foreign-language words,
does not induce this same alignment (Figure~\ref{fig:retrieval-all8}, Appendix~\ref{app:retrieval}), suggesting that grammatically and semantically coherent code-switching is crucial.

\paragraph{Cross-lingual alignment persists through non-CS training.}
We measure bitext retrieval precision across model checkpoints (Figure~\ref{fig:retrievaltraj}). Cross-lingual alignment improves during the word-level and sentence-level CS stages and is maintained even through the non-CS training stage.

\section{Representations of Code-Switched Words}

\label{sec:words}

Does CS training improve the representations of only the words seen in embedded contexts,
or does the cross-lingual alignment extend across the model's entire vocabulary?

\paragraph{Embedded vs.\ never-embedded words.}
We call a word \emph{embedded} if it was ever inserted as a foreign-language word during word-level CS,
and \emph{never-embedded} if it was neither embedded nor replaced by an embedded word.
We exclude replaced words since they have systematically lower frequency in the CS corpus due to being replaced.
We compute a word $w$'s representation as follows.
We first collect a set of sentence contexts from the original corpus that contain $w$.
We then compute a forward pass and take the mean of the residual stream vectors across the token positions that constitute $w$, across the set of sentences, and across layers $6$--$11$.

\paragraph{Median percentile rank metric.}
Similarly to the sentence-level bitext retrieval metric (\S\ref{sec:mechanism}),
for each word $w^A$ in language A,
we construct a set of words $w^B_1, \ldots w^B_N$ in language B containing the translation $w^B_*$ of $w^A$.
We then rank the set according to the same CSLS-adjusted cosine similarity metric,
with regards to the word representations described above,
and record the \textbf{percentile rank} of $w^B_*$.
(This contrasts with the precision@1 metric where we would only check if $w^B_*$ is the most similar word to $w^A$.)
We compute these percentile ranks across $N=1{,}102$ embedded/non-embedded word pairs and,
for each of the $4$ language directions for which we have an openly accessible bilingual dictionary,
plot the median in Figure~\ref{fig:exposuregen}.
See Appendix~\ref{app:wordlevel} for details on choosing the candidate sets and labeling translation pairs.

\begin{figure}[t]
\centering
\includegraphics[width=\columnwidth]{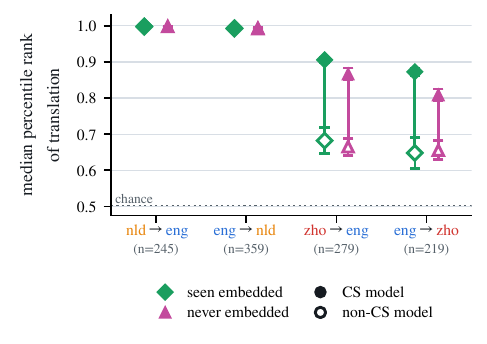}
\caption{
Median percentile rank of a word's correct translation among a set of target-language words matched by part of speech and frequency (Appendix~\ref{app:wordlevel}). The number of word pairs for each language pair is shown in parentheses.}
\label{fig:exposuregen}
\end{figure}

\paragraph{Never-embedded words also become aligned by CS training.}
Relative to the non-CS baseline, CS training raises the median percentile rank of the translation by $+0.064$ for embedded words and $+0.045$ for never-embedded words (Figure~\ref{fig:exposuregen}).
The English--Dutch translations are already closely aligned in both the CS and non-CS models. Here in word-level alignment, as in sentence-level alignment (\S\ref{sec:mechanism}), we see that CS training aligns representations especially across distinct scripts.

\section{Conclusion}
\label{sec:conclusion}
We train and analyze small autoregressive language models on code-switched curriculum learning under BabyLM competition constraints (10 epochs on 100M words shared between Dutch, English, and Chinese).
When trained in a three-stage learning curriculum,
from word-level code-switching to sentence-level switching to monolingual documents,
a model trained on code-switched data achieves a $0.36$-point gain on the BabyLM evaluation suite,
on average across eight seeds,
relative to a baseline trained on a matched corpus of monolingual documents (RQ1, \S\ref{sec:effect}).
However, under a randomly shuffled data ordering,
there is no statistically significant improvement.
We show that code-switching in pretraining aligns the representations of parallel sentences (RQ2, \S\ref{sec:mechanism})
and additionally show that embedded words become more closely aligned to their translations than words never seen embedded in a code-switched context (RQ3, \S\ref{sec:words}).
We release our corpora and training code to support future research on multilingual data composition and data ordering at a fixed data budget.

\section*{Limitations}

\label{sec:limitations}

\paragraph{Language coverage.}
We study only trilingual models in English, Dutch, and Chinese so as to match the BabyLM challenge evaluation suite.
Whether the curriculum and the alignment mechanism generalize to more distant or lower-resource pairs, or to more than three languages at once, is untested.

\paragraph{LLM-generated code-switching.}
Our LLM-driven synthetic CS corpus generation method would be expensive to scale to larger datasets.
It also results in a confound whereby the sentence-level CS data and the document translation control corpus risk having a different text distribution than BabyBabelLM,
which might thus be responsible for performance improvements rather than the code-switching intervention itself.

\paragraph{Model and budget scope.}
All of our results are for a single model size and architecture (${\sim}98$M-parameter GPT-2 at the $100$M-word budget).
Future work could investigate whether the benefits of training on code-switched text persist in larger models and data budgets.

\section*{Ethics Statement}

\paragraph{Data provenance and released artifacts.}
All training text derives from the BabyBabelLM corpora \citep{jumelet2026babybabellm},
distributed under the BabyLM community's access terms.
We did not collect data from human subjects;
manual review of generated samples was done by the joint first authors.
Our models are ${\sim}98$M-parameter research artifacts with no instruction tuning or safety alignment.
We release them and our corpora under the MIT license.

\paragraph{Compute.}
The experiments in this paper required training $64$ models, each on a single NVIDIA H100 80GB GPU.
Pretraining one model takes ${\sim}2.3$ GPU-hours and evaluating a model takes an additional ${\sim}0.8$.
This adds up to roughly $200$ GPU-hours in total,
excluding the LLM inference used to synthesize the corpora,
the checkpoint analyses of \S\ref{sec:mechanism}--\S\ref{sec:words}, and re-run jobs.

\section*{Acknowledgments}
We thank Nihal Nayak, Sara Kangaslahti, Greta Tuckute, and the Harvard ML Foundations Group for conversations regarding multilingual models.
AC is grateful to be funded by the Kempner Graduate Fellowship.
Computation was performed on the Kempner Institute's H100 cluster at Harvard University.
We wrote most of our analysis and plotting code with help from Anthropic's Claude LLM.

\bibliography{custom}  

\appendix
\setcounter{dbltopnumber}{3}
\setlength{\textfloatsep}{10pt plus 2pt minus 2pt}
\setlength{\floatsep}{8pt plus 2pt minus 2pt}
\setlength{\dbltextfloatsep}{10pt plus 2pt minus 2pt}
\setlength{\dblfloatsep}{8pt plus 2pt minus 2pt}
\renewcommand{\textfraction}{0.1}
\makeatletter
\setlength{\@dblfptop}{0pt}\setlength{\@dblfpsep}{10pt plus 2pt}\setlength{\@dblfpbot}{0pt plus 1fil}
\setlength{\@fptop}{0pt}\setlength{\@fpsep}{10pt plus 2pt}\setlength{\@fpbot}{0pt plus 1fil}
\makeatother
\setlength{\abovecaptionskip}{5pt}
\setcounter{bottomnumber}{1}
\setcounter{totalnumber}{2}
\setcounter{topnumber}{2}
\setcounter{dbltopnumber}{2}
\setcounter{totalnumber}{2}

\section{Data Generation Details}
\label{app:data}

\paragraph{Sources, budget, units.}
We download the BabyBabelLM \citep{jumelet2026babybabellm} datasets from HuggingFace (IDs \texttt{BabyLM-community/babylm-\{eng,nld,zho\}}) at revisions
(\texttt{b78a9336}, \texttt{1aa063f7}, \texttt{600a6657}) respectively.
We normalize according to the following per-language byte premiums \citep{arnett2024bytepremium}:
English $1.0$, Dutch $1.0516$, Simplified Chinese $0.935966$.
The word unit is the
\texttt{num-tokens} column of the released datasets.
For English and Dutch this column counts whitespace tokens. For Chinese, per the \texttt{babylm-zho} dataset card, it is the subword count under the Qwen3-0.6B tokenizer (we verified an exact match on a sample of documents).
We approximate the Chinese count as $0.711$ units per character, the corpus-level ratio of \texttt{num-tokens} to characters measured on \texttt{babylm-zho}.
We split each source document into passages of about $300$ words at sentence boundaries.
Chinese passages in which Han characters make up fewer than $50\%$ of Han-plus-Latin characters are
dropped as romanized-pinyin transcripts (the Chinese portion of CHILDES \citep{macwhinney2000childes}), as are passages
under $15$ words in any language.
\begin{figure*}[t]
\centering
\small
\setlength{\tabcolsep}{6pt}\renewcommand{\arraystretch}{1.25}
\begin{tabular}{@{}l >{\raggedright\arraybackslash}p{4.55cm} >{\raggedright\arraybackslash}p{4.55cm} >{\raggedright\arraybackslash}p{4.55cm}@{}}
\toprule
 & \textbf{\textcolor{intracol}{Word-level CS}}
 & \textbf{\textcolor{sentcol}{Sentence-level CS}}
 & \textbf{\textcolor{monocol}{Non-CS}} \\
\midrule
\textcolor{langEN}{\textbf{English}}
 & \dtag{EN\,$\leftarrow$\,NL}\newline Rondo is a \embNL{dorp} \embNL{in de VS staat} Arkansas.
 & \dtag{EN\,$\leftrightarrow$\,ZH}\newline Landaff is a district in Wales.\ \embZH{\zh{它是英国作家罗尔德·达尔的出生地。}}
 & Rondo is a town in the US state of Arkansas. \\
\addlinespace[5pt]
\textcolor{langNL}{\textbf{Dutch}}
 & \dtag{NL\,$\leftarrow$\,ZH}\newline `Ga \embZH{\zh{走}}, voordat het te laat \embZH{\zh{太晚}} is!' snauwde opa \embZH{\zh{爷爷}}.
 & \dtag{NL\,$\leftrightarrow$\,EN}\newline Iemand voor schut zetten. \embEN{To put someone to shame is to insult someone.}
 & `Ga, voordat het te laat is!' snauwde opa. \\
\addlinespace[5pt]
\textcolor{langZH}{\textbf{Chinese}}
 & \dtag{ZH\,$\leftarrow$\,EN}\newline \zh{但是我不会}\,\embEN{swim}\,\zh{呀,你那}\,\embEN{swimming posture}\,\zh{也分解不对呀。}
 & \dtag{ZH\,$\leftrightarrow$\,NL}\newline \zh{我知道了。} \embNL{Dan moeten we morgen zeker ijs eten.}
 & \zh{但是我不会游泳呀,你那游泳姿势也分解不对呀。} \\
\bottomrule
\end{tabular}
\caption{\textbf{Verbatim corpus examples} by matrix language (rows) and switch type (columns).
Embedded-language spans are colored by language (\embEN{English}, \embNL{Dutch}, \embZH{Chinese}).
The non-CS cell in each row is the true source document of that row's word-level example.}
\label{fig:composition}
\end{figure*}

\paragraph{Generation, batching, cost.}
Here we describe our final production system to generate of the word- and sentence-level CS corpus. It uses \texttt{deepseek-v4-flash} via OpenRouter.
Each request carries eight passages sharing one (matrix, embedded) language pair as JSON lines and returns one JSON line per input id.
The full translations used only by the document translation control (\S\ref{sec:data}) were generated with \texttt{claude-haiku-4-5} via the Anthropic Message Batches API, with the prompt of Appendix~\ref{app:prompts}.
Each switch type covers the same $413{,}302$ passages ($51{,}666$ requests); the total API cost was ${\approx}\$284$.

\paragraph{Quality verification.}
Every generation must contain both the matrix and the embedded language.
We check language presence with (\texttt{lingua}):
Chinese counts as present at ${\ge}3$ Han characters;
in Latin-script text a language counts as present when it covers ${\ge}8\%$ of the detected span length.
We reject outputs that are identical to the source (no switching),
that miss the matrix language (a full translation),
or that miss the embedded language,
and we reject outputs whose character count falls outside $0.25\times$--$4\times$ that of the
source, to catch truncations and runaway generations.
Acceptance rates, as a fraction of requested passages, are $0.78$--$0.89$ for word-level switching in
every direction and $0.79$--$0.87$ for sentence-level switching into an English or Dutch matrix,
but fall to $0.47$--$0.56$ for sentence-level switching into a Chinese matrix, where most
rejections are unchanged or fully translated outputs.
In total ${\sim}348$K word-level and ${\sim}323$K sentence-level passages make it past this initial quality filter.

\paragraph{Document reassembly.}
Accepted passages are stitched back into documents in order. A document whose every passage was rejected is dropped, and one with a rejected passage is reassembled with that passage missing ($3.7\%$ of CS documents, $1.8\%$ with a gap in the middle of the passage).
The non-CS baseline document is stitched from the source text of the same accepted passages, so both the CS and non-CS versions have identical gaps.

\paragraph{Budget accounting.}
The word-level, sentence-level, and non-CS thirds occupy disjoint document ids, so that no same document is used for two different thirds of our corpus.
Because we found that per-language word heuristics do not work well on mixed text,
we check compliance to the 100M adjusted-word budget twice. We use a script-aware word count (Han characters ${\times}0.711$ plus whitespace tokens of the Latin remainder), and we also count the total UTF-8 bytes ($543$~MB${\,\equiv\,}100$M English words).
By both metrics, the CS corpus is within budget (${\approx}96.6$M words / $485$~MB) and near-identical in volume to its non-CS baseline ($95.7$M words / $482$~MB); the $99.6$M of Table~\ref{tab:composition} counts the same documents with the per-language heuristic above.
Under our $16{,}384$-entry BPE vocabulary the CS corpus is $138.9$M tokens and the non-CS corpus $135.0$M, a $2.9\%$ difference;
the two code-switched thirds tokenize to $2$--$12\%$ more tokens than their source documents, most for Chinese-matrix text with translated Latin-script sentences.

\paragraph{Examples from control corpora.}
The following two examples come from the word salad corpus (\S\ref{sec:data}):
\begin{itemize}\setlength{\itemsep}{2pt}\setlength{\parskip}{0pt}
  \item[] \dtag{EN\,$\leftarrow$\,NL}~Transduction is a \embNL{wisselkoersen} term.
  It can mean:
  \item[] \dtag{ZH\,$\leftarrow$\,NL}~\zh{曾以“女排精神”引}\embNL{mevrouw}\zh{领一代风骚的中国}\embNL{voelen}\zh{女排，今}\embNL{gravin}\zh{天再度擎起世界冠军大旗，怎能不让国人惊喜？}
\end{itemize}
The following two come from the document translation corpus:
\begin{itemize}\setlength{\itemsep}{2pt}\setlength{\parskip}{0pt}
  \item[] \dtag{EN\,$\to$\,NL}~Burien is a city in King County, Washington, United States.
  $\Rightarrow$ \embNL{Burien is een stad in King County, Washington, Verenigde Staten.}
  \item[] \dtag{ZH\,$\to$\,EN}~\zh{不然它们就吃撑了。} $\Rightarrow$ \embEN{Otherwise they will eat until they are stuffed.}
\end{itemize}

\section{Generation Model and Prompts}
\label{app:prompts}

We chose the generation model through a couple different steps. First we scored a $36$-passage sample (three per matrix$\times$embedded$\times$switch-type cell) from several frontier LLMs (as of June 2026) with the quality gate of Appendix~\ref{app:data}, supplemented by manual review from native English--Dutch and English--Chinese speakers.
Pass rates were DeepSeek-V4-Flash \citep{deepseek2026v4} $0.72$, \texttt{claude-haiku-4-5} $0.67$, Gemini~2.5~Flash-Lite $0.33$, and Qwen3-30B-A3B $0.19$. The Qwen3 models  often drifted into full translation on cross-script pairs, whereas Haiku often failed to produce embedded Chinese words. We therefore used \texttt{deepseek-v4-flash} for the full run, which from manual inspection gave good results for each kind of code-switching.
Unlike \citet{yoo2025cscl}, we did not need to supply parallel translations to the generator to obtain reasonable CS passages.

Our generation prompts follow verbatim.
\texttt{\{matrix\}}/\texttt{\{embedded\}} refer to the language names (English/Dutch/Chinese (Simplified)).
Each prompt is followed by the instruction to return exactly one JSON line per input id with the same ids,
and then by the input passages as JSON lines of the form \texttt{\{"id": ..., "text": ...\}}.

\paragraph{Word-level CS.}
{\scriptsize\begin{verbatim}
Task: rewrite each {matrix} text as
INTRASENTENTIAL code-switching with {embedded}.
Rules:
- Keep {matrix} as the matrix language: it
  carries the grammar, word order, and most
  function words. Most words stay in {matrix}.
- CRITICAL: do NOT translate sentences fully
  into {embedded}. {matrix} must remain
  dominant throughout.
- Embed natural {embedded} content words and
  short phrases INSIDE sentences (nouns, verbs,
  adjectives, technical terms, short NPs).
- Switch only at grammatically permissible
  boundaries. Do not switch inside a word.
- Target roughly 25-45% of content words/
  phrases in {embedded}; every non-trivial
  sentence should contain >=1 {embedded} switch.
- Preserve meaning exactly. Do not add, drop,
  summarise, or invent facts.
- Do NOT translate a whole sentence into
  {embedded}: mix both languages within
  sentences.
- Use the native script for {embedded}.
- Preserve sentence order and any list/markup.
\end{verbatim}}

\paragraph{Sentence-level CS.}
{\scriptsize\begin{verbatim}
Task: rewrite each {matrix} text as
SENTENCE-LEVEL code-switching with {embedded}.
Rules:
- Keep the original sentence ORDER. Translate
  roughly half of the sentences fully into
  {embedded}; leave the rest in {matrix}.
- CRITICAL: do NOT translate the whole passage.
  About half of sentences must remain {matrix}.
  Output with no {matrix} sentences is invalid.
- Each sentence stays internally monolingual:
  all {matrix} or all {embedded}. Do NOT mix
  languages inside a sentence.
- Translate faithfully; preserve meaning. Do
  not add, drop, or invent facts.
- Leave short fragments, list items, headings,
  or markup-like lines unchanged if translating
  would corrupt them.
- Use the native script for {embedded}.
\end{verbatim}}

\paragraph{Document translation control.}
{\scriptsize\begin{verbatim}
Task: translate each {matrix} passage FAITHFULLY
and COMPLETELY into {embedded}.
Rules:
- Translate the ENTIRE passage into {embedded};
  no {matrix} words should remain (proper nouns
  may stay).
- Preserve meaning, sentence order, and
  structure exactly; do not add, drop, or
  summarise.
- Natural, fluent {embedded}; not word-for-word
  glossing.
- Use the native script for {embedded}.
\end{verbatim}}

\section{Evaluation Suite}
\label{app:evalsuite}

\begin{table*}[p]
  \centering \footnotesize
  \setlength{\tabcolsep}{4pt}\renewcommand{\arraystretch}{1.15}
  \resizebox{\textwidth}{!}{%
  \begin{tabular}{@{}l c *{8}{c}@{}}
    \toprule
    Task & Chance & Shuf & CS/Shuf & Curr & CS/Curr & Word & Sent & Par & Salad \\
    \midrule
    \multicolumn{10}{@{}l}{\textcolor{langEN}{\emph{English}}} \\
    \quad BLiMP      & 50.0 & 67.00$\pm$0.97 & 68.27$\pm$0.56 & 66.17$\pm$0.90 & 67.18$\pm$0.98 & 67.95$\pm$0.45 & 68.68$\pm$0.62 & 68.03$\pm$0.73 & 65.88$\pm$0.78 \\
    \quad MultiBLiMP & 50.0 & 83.75$\pm$1.23 & 85.70$\pm$1.55 & 83.64$\pm$0.84 & 84.74$\pm$0.85 & 83.85$\pm$1.27 & 85.55$\pm$0.89 & 85.50$\pm$1.22 & 82.48$\pm$1.66 \\
    \quad HellaSwag  & 25.0 & 26.67$\pm$0.14 & 26.48$\pm$0.14 & 26.60$\pm$0.10 & 26.56$\pm$0.26 & 26.63$\pm$0.14 & 26.59$\pm$0.24 & 26.59$\pm$0.17 & 26.52$\pm$0.16 \\
    \quad Winogrande & 50.0 & 50.37$\pm$0.66 & 50.06$\pm$1.08 & 50.30$\pm$1.06 & 50.60$\pm$0.75 & 49.97$\pm$0.63 & 50.01$\pm$0.62 & 50.13$\pm$0.80 & 49.89$\pm$0.51 \\
    \quad XStoryCloze& 50.0 & 50.60$\pm$0.60 & 50.22$\pm$0.71 & 50.24$\pm$0.48 & 50.24$\pm$0.37 & 50.18$\pm$0.57 & 49.63$\pm$0.72 & 50.74$\pm$0.62 & 50.07$\pm$0.76 \\
    \quad Global PIQA& 25.0 & 38.49$\pm$2.20 & 37.83$\pm$1.47 & 37.57$\pm$1.68 & 37.53$\pm$2.07 & 38.08$\pm$1.47 & 37.34$\pm$2.84 & 38.13$\pm$1.88 & 38.38$\pm$2.01 \\
    \quad ARC        & 25.0 & 26.09$\pm$2.47 & 24.66$\pm$2.56 & 26.07$\pm$1.33 & 26.20$\pm$0.81 & 25.57$\pm$1.57 & 26.25$\pm$1.74 & 25.76$\pm$1.21 & 25.57$\pm$1.59 \\
    \quad Belebele   & 25.0 & 22.44$\pm$2.04 & 21.59$\pm$2.17 & 21.88$\pm$2.45 & 22.02$\pm$1.94 & 22.87$\pm$1.91 & 22.87$\pm$2.23 & 20.88$\pm$1.76 & 23.65$\pm$2.10 \\
    \quad BMLAMA     & \textemdash & 23.58$\pm$3.87 & 21.29$\pm$4.00 & 22.79$\pm$1.91 & 22.15$\pm$3.57 & 23.83$\pm$2.13 & 23.02$\pm$1.61 & 21.99$\pm$3.72 & 20.79$\pm$3.99 \\
    \quad MNLI       & 33.3 & 45.07$\pm$1.74 & 46.50$\pm$1.39 & 45.09$\pm$0.84 & 47.02$\pm$2.43 & 46.57$\pm$1.99 & 45.43$\pm$1.99 & 46.23$\pm$1.45 & 45.22$\pm$1.19 \\
    \quad SIB-200    & 14.3 & 72.00$\pm$3.79 & 72.25$\pm$2.22 & 71.38$\pm$2.63 & 73.63$\pm$1.77 & 73.31$\pm$2.90 & 71.31$\pm$2.42 & 71.75$\pm$2.28 & 72.12$\pm$2.33 \\
    \quad TruthfulQA & \textemdash & 24.00$\pm$3.77 & 23.77$\pm$6.17 & 25.78$\pm$3.55 & 27.12$\pm$3.69 & 27.01$\pm$4.26 & 28.24$\pm$4.42 & 27.57$\pm$5.17 & 28.57$\pm$5.10 \\
    \quad XNLI       & 33.3 & 43.26$\pm$0.37 & 44.62$\pm$1.20 & 43.71$\pm$1.36 & 44.26$\pm$1.18 & 43.72$\pm$1.03 & 44.33$\pm$1.24 & 44.27$\pm$0.83 & 44.39$\pm$0.92 \\
    \quad POS        & \textemdash & 92.79$\pm$0.29 & 92.98$\pm$0.30 & 92.75$\pm$0.26 & 92.76$\pm$0.13 & 92.90$\pm$0.28 & 92.79$\pm$0.23 & 92.84$\pm$0.17 & 92.92$\pm$0.09 \\
    \midrule
    \multicolumn{10}{@{}l}{\textcolor{langNL}{\emph{Dutch}}} \\
    \quad BLiMP-NL   & 50.0 & 77.35$\pm$0.55 & 77.36$\pm$0.63 & 76.84$\pm$0.71 & 76.80$\pm$0.62 & 77.49$\pm$0.44 & 77.57$\pm$0.46 & 77.27$\pm$0.52 & 76.15$\pm$0.79 \\
    \quad MultiBLiMP & 50.0 & 90.06$\pm$0.56 & 90.41$\pm$0.30 & 89.40$\pm$0.31 & 89.60$\pm$0.62 & 90.19$\pm$0.56 & 90.28$\pm$0.67 & 90.27$\pm$0.60 & 89.16$\pm$0.59 \\
    \quad HellaSwag  & 25.0 & 26.40$\pm$0.20 & 26.43$\pm$0.13 & 26.47$\pm$0.12 & 26.32$\pm$0.20 & 26.37$\pm$0.15 & 26.38$\pm$0.18 & 26.53$\pm$0.13 & 26.25$\pm$0.22 \\
    \quad Winogrande & 50.0 & 49.77$\pm$0.81 & 50.59$\pm$1.21 & 49.33$\pm$0.61 & 49.79$\pm$0.52 & 49.95$\pm$0.62 & 49.81$\pm$0.45 & 49.38$\pm$0.91 & 49.99$\pm$1.37 \\
    \quad XCOMPS     & 50.0 & 52.90$\pm$0.58 & 52.34$\pm$0.43 & 52.64$\pm$0.69 & 52.40$\pm$0.31 & 52.35$\pm$0.52 & 52.42$\pm$0.56 & 52.53$\pm$0.85 & 52.28$\pm$0.43 \\
    \quad XStoryCloze& 50.0 & 48.06$\pm$0.49 & 49.02$\pm$0.54 & 48.57$\pm$0.37 & 48.46$\pm$0.58 & 48.39$\pm$0.68 & 48.82$\pm$0.40 & 48.36$\pm$0.48 & 48.12$\pm$0.28 \\
    \quad Global PIQA& 25.0 & 39.25$\pm$1.74 & 37.61$\pm$1.81 & 39.64$\pm$1.90 & 39.22$\pm$1.00 & 36.63$\pm$1.69 & 38.35$\pm$2.49 & 39.34$\pm$0.87 & 38.84$\pm$2.27 \\
    \quad ARC        & 25.0 & 26.67$\pm$1.28 & 26.35$\pm$1.40 & 26.25$\pm$2.18 & 26.46$\pm$1.81 & 26.54$\pm$2.25 & 26.07$\pm$2.61 & 24.90$\pm$1.72 & 26.30$\pm$1.71 \\
    \quad Belebele   & 25.0 & 24.01$\pm$1.60 & 23.22$\pm$1.62 & 22.73$\pm$1.58 & 23.86$\pm$1.39 & 21.24$\pm$1.49 & 22.59$\pm$2.12 & 24.22$\pm$1.80 & 23.22$\pm$2.05 \\
    \quad BMLAMA     & \textemdash & 22.88$\pm$2.92 & 22.12$\pm$2.24 & 23.12$\pm$2.71 & 22.87$\pm$1.87 & 21.78$\pm$4.19 & 22.55$\pm$2.51 & 22.08$\pm$3.56 & 19.42$\pm$4.61 \\
    \quad INCLUDE    & 25.0 & 31.58$\pm$3.13 & 31.47$\pm$1.49 & 30.47$\pm$4.95 & 31.47$\pm$2.65 & 30.13$\pm$5.00 & 32.81$\pm$2.51 & 32.81$\pm$5.18 & 30.58$\pm$4.62 \\
    \quad MNLI       & 33.3 & 46.66$\pm$1.26 & 46.59$\pm$1.71 & 47.22$\pm$0.87 & 46.67$\pm$1.61 & 47.71$\pm$1.15 & 46.99$\pm$1.71 & 47.22$\pm$1.22 & 46.23$\pm$1.52 \\
    \quad SIB-200    & 14.3 & 67.69$\pm$3.24 & 69.50$\pm$2.04 & 69.06$\pm$2.26 & 70.62$\pm$2.50 & 68.81$\pm$1.56 & 68.44$\pm$2.15 & 69.50$\pm$1.77 & 68.00$\pm$2.59 \\
    \quad TruthfulQA & \textemdash & 22.77$\pm$3.94 & 24.44$\pm$3.13 & 23.55$\pm$2.61 & 23.55$\pm$3.13 & 23.77$\pm$3.88 & 24.11$\pm$3.31 & 24.78$\pm$2.69 & 23.10$\pm$2.76 \\
    \quad POS        & \textemdash & 94.45$\pm$0.30 & 94.52$\pm$0.18 & 94.28$\pm$0.20 & 94.24$\pm$0.33 & 94.48$\pm$0.16 & 94.33$\pm$0.15 & 94.54$\pm$0.15 & 94.41$\pm$0.21 \\
    \midrule
    \multicolumn{10}{@{}l}{\textcolor{langZH}{\emph{Chinese}}} \\
    \quad ZhoBLiMP   & 50.0 & 75.95$\pm$0.91 & 76.88$\pm$0.81 & 75.52$\pm$1.64 & 74.84$\pm$0.39 & 74.65$\pm$1.21 & 76.28$\pm$0.52 & 75.22$\pm$0.94 & 72.48$\pm$0.97 \\
    \quad HellaSwag  & 25.0 & 26.91$\pm$0.12 & 26.41$\pm$0.17 & 26.86$\pm$0.14 & 26.63$\pm$0.21 & 26.83$\pm$0.09 & 26.50$\pm$0.11 & 26.64$\pm$0.08 & 26.93$\pm$0.05 \\
    \quad Winogrande & 50.0 & 48.75$\pm$0.84 & 49.45$\pm$0.94 & 49.40$\pm$0.63 & 49.34$\pm$0.86 & 49.61$\pm$0.78 & 49.88$\pm$0.61 & 49.48$\pm$0.27 & 49.20$\pm$0.95 \\
    \quad XCOMPS     & 50.0 & 53.71$\pm$0.56 & 53.99$\pm$0.57 & 53.55$\pm$0.24 & 53.73$\pm$0.32 & 53.87$\pm$0.43 & 53.81$\pm$0.56 & 53.62$\pm$0.42 & 51.17$\pm$0.48 \\
    \quad XStoryCloze& 50.0 & 47.93$\pm$0.54 & 48.60$\pm$0.54 & 47.42$\pm$0.53 & 48.42$\pm$0.42 & 49.03$\pm$0.52 & 49.05$\pm$0.65 & 48.90$\pm$0.44 & 47.92$\pm$0.50 \\
    \quad Global PIQA& 25.0 & 32.77$\pm$1.65 & 32.13$\pm$2.15 & 33.35$\pm$1.45 & 33.22$\pm$1.28 & 32.66$\pm$1.64 & 33.05$\pm$2.90 & 31.12$\pm$1.58 & 31.95$\pm$1.80 \\
    \quad ARC        & 25.0 & 28.02$\pm$2.79 & 26.43$\pm$1.22 & 26.80$\pm$1.54 & 26.77$\pm$2.11 & 26.56$\pm$1.59 & 26.82$\pm$2.32 & 25.49$\pm$1.02 & 26.17$\pm$1.30 \\
    \quad Belebele   & 25.0 & 20.95$\pm$1.96 & 21.16$\pm$1.54 & 22.23$\pm$1.34 & 22.73$\pm$2.31 & 22.94$\pm$1.94 & 21.31$\pm$1.05 & 22.80$\pm$2.11 & 23.08$\pm$3.30 \\
    \quad BMLAMA     & \textemdash & 18.34$\pm$2.56 & 16.20$\pm$4.07 & 17.42$\pm$2.66 & 15.90$\pm$3.86 & 17.28$\pm$2.59 & 18.49$\pm$2.56 & 15.98$\pm$3.67 & 15.21$\pm$2.92 \\
    \quad INCLUDE    & 25.0 & 26.45$\pm$3.44 & 26.45$\pm$3.99 & 25.11$\pm$3.95 & 26.79$\pm$2.70 & 27.12$\pm$3.88 & 25.56$\pm$2.08 & 24.33$\pm$3.96 & 25.67$\pm$4.32 \\
    \quad MNLI       & 33.3 & 44.64$\pm$1.67 & 47.43$\pm$2.33 & 45.45$\pm$2.12 & 47.42$\pm$2.58 & 45.21$\pm$1.83 & 45.62$\pm$1.47 & 46.38$\pm$3.02 & 46.00$\pm$1.29 \\
    \quad SIB-200    & 14.3 & 75.88$\pm$2.46 & 78.25$\pm$2.67 & 76.56$\pm$1.80 & 78.56$\pm$1.40 & 78.19$\pm$2.58 & 77.81$\pm$1.39 & 77.00$\pm$2.24 & 77.13$\pm$1.94 \\
    \quad TruthfulQA & \textemdash & 22.99$\pm$2.85 & 21.99$\pm$4.32 & 23.33$\pm$2.31 & 22.10$\pm$3.43 & 22.32$\pm$1.58 & 23.21$\pm$2.29 & 24.11$\pm$2.53 & 23.88$\pm$2.33 \\
    \quad XNLI       & 33.3 & 42.66$\pm$1.01 & 42.86$\pm$0.47 & 42.47$\pm$0.43 & 43.67$\pm$1.54 & 43.73$\pm$1.09 & 42.83$\pm$0.71 & 41.83$\pm$1.50 & 40.47$\pm$2.77 \\
    \quad POS        & \textemdash & 89.67$\pm$0.22 & 90.56$\pm$0.31 & 89.81$\pm$0.22 & 90.05$\pm$0.29 & 90.26$\pm$0.27 & 89.87$\pm$0.27 & 90.11$\pm$0.38 & 90.02$\pm$0.19 \\
    \midrule
    \emph{Leaderboard} & & 46.44$\pm$0.25 & 46.55$\pm$0.47 & 46.36$\pm$0.37 & 46.72$\pm$0.25 & 46.59$\pm$0.24 & 46.70$\pm$0.29 & 46.56$\pm$0.38 & 46.08$\pm$0.34 \\
    \bottomrule
  \end{tabular}}
  \caption{Accuracy on the BabyLM evaluation suite per task and per language. We report (\%) mean\,$\pm$\,SD across eight seeds per condition.
  Chance performance is random-guess accuracy ($1/\#$options) where applicable.
  Word/Sent/Par/Salad are the shuffled-ordering controls (\S\ref{sec:data}).
  \emph{Leaderboard} is the official multilingual-average score as in Table~\ref{tab:results}.}
  \label{tab:sweep}
\end{table*}

\begin{table}[tp]
  \centering \small
  \setlength{\tabcolsep}{3pt}
  \begin{tabular}{@{}l ccc@{}}
    \toprule
    Condition & English & Dutch & Chinese \\
    \midrule
    Shuf & 1.60$\pm$0.05 & 1.56$\pm$0.04 & 1.78$\pm$0.03 \\
    CS/Shuf & 1.45$\pm$0.01 & 1.45$\pm$0.01 & 1.75$\pm$0.01 \\
    Curr & 1.56$\pm$0.02 & 1.53$\pm$0.01 & 1.78$\pm$0.01 \\
    CS/Curr & 1.56$\pm$0.03 & 1.56$\pm$0.03 & 1.80$\pm$0.01 \\
    \midrule
    Word-level CS only & 1.52$\pm$0.04 & 1.53$\pm$0.03 & 1.78$\pm$0.02 \\
    Sentence-level CS only & 1.42$\pm$0.02 & 1.42$\pm$0.02 & 1.73$\pm$0.02 \\
    Document translation & 1.63$\pm$0.06 & 1.61$\pm$0.05 & 1.78$\pm$0.03 \\
    Word salad & 1.65$\pm$0.03 & 1.61$\pm$0.03 & 1.79$\pm$0.01 \\
    \bottomrule
  \end{tabular}
  \caption{Bits per UTF-8 byte on $1{,}200$ held-out BabyBabelLM documents per language (lower is better). The first $512$ tokens of each document are scored.}
  \label{tab:heldout}
\end{table}

We evaluate with the official BabyLM 2026 multilingual evaluation pipeline \citep{choshen2026babylm},
which scores each task in each of the three languages where available.

\paragraph{Zero-shot tasks.}
The zero-shot tasks consist of minimal-pair and multiple-choice tasks scored by comparing candidate log-probabilities.
Grammar evaluations include BabyLM's filtered subset of BLiMP \citep{warstadt2020blimp} for English, with BLiMP-NL \citep{suijkerbuijk2025blimpnl} and ZhoBLiMP \citep{liu2026zhoblimp} as its Dutch and Chinese counterparts, and MultiBLiMP \citep{jumelet2026multiblimp} in English and Dutch.
Common-sense semantic reasoning benchmarks include HellaSwag \citep{zellers2019hellaswag};
Winogrande \citep{sakaguchi2020winogrande};
XStoryCloze \citep{lin2022xglm}; XCOMPS \citep{he2025xcomps}; Global PIQA \citep{chang2025globalpiqa}; and the Chinese Hanzi Structure and Hanzi Pinyin minimal-pair tasks distributed with the pipeline.
The latter two are hidden tasks:
the leaderboard computes their scores server-side and excludes them from every average,
so they do not enter the leaderboard score or Table~\ref{tab:sweep}.

\paragraph{Fine-tuning tasks.}
ARC \citep{clark2018arc}; Belebele \citep{bandarkar2024belebele}; BMLAMA
\citep{qi2023bmlama}; MNLI \citep{williams2018mnli}; SIB-200 \citep{adelani2024sib200};
TruthfulQA \citep{lin2022truthfulqa}; XNLI \citep{conneau2018xnli}; INCLUDE
\citep{romanou2025include}; and cross-lingual POS tagging over Universal Dependencies
treebanks \citep{demarneffe2021ud}.
Each classification task is fine-tuned and evaluated within one language
(up to $10$ epochs with early stopping at patience $3$, learning rate $5\times10^{-5}$, batch size $64$, one fine-tuning seed); only POS tagging is trained jointly on all three languages.

\paragraph{Harness details.}
We run the harness with the organizers' fix for the beginning-of-sequence (BOS) token: the harness recovers the continuation
tokens as the suffix of the tokenized prompt$+$continuation, which breaks for tokenizers that, like
the baseline's and ours, wrap every string in \texttt{<s>}\ldots\texttt{</s>}; the fix tokenizes without
special tokens and prepends a single \texttt{<s>}.

\section{Details on Bitext Retrieval}
\label{app:retrieval}

\begin{figure*}[tp]
\centering
\includegraphics[width=0.9\textwidth]{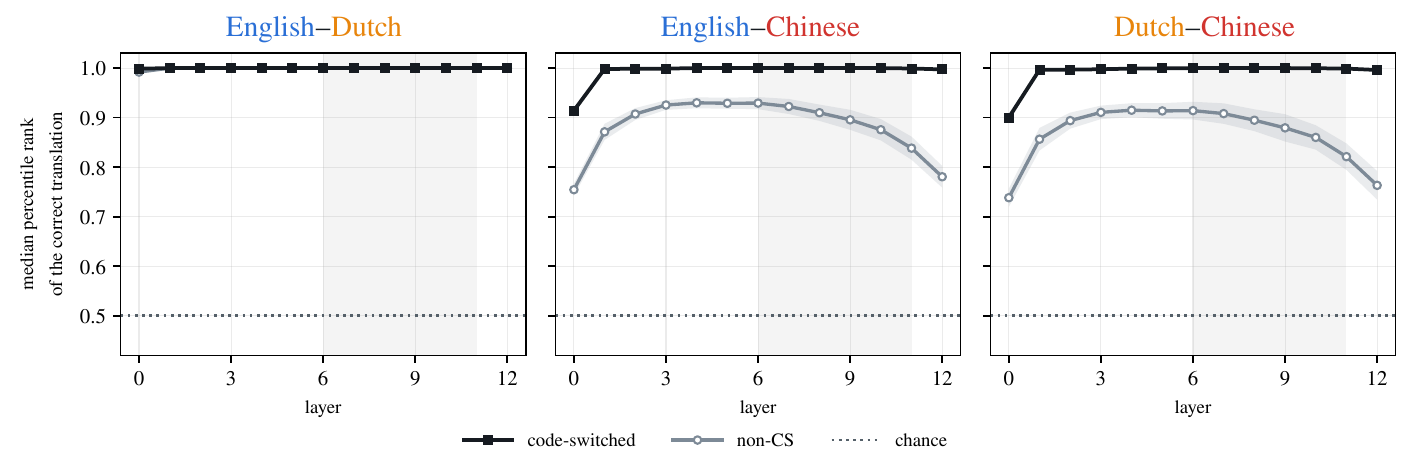}
\caption{In addition to sentence-level bitext retrieval precision@1 (Figure~\ref{fig:retrieval}),
here we show the per-layer median percentile rank of the correct translation.
Averaged over layers 6--11, the non-CS models place the correct translation at median rank $105$ and $121$ out of $997$ for English-Chinese and Dutch-Chinese respectively,
far above chance at $499$.
This suggests that code-switching does not create alignment ``from scratch'',
but rather ``sharpens'' a general degree of alignment found even in the non-CS models.}
\label{fig:retrieval-medrank}
\end{figure*}

\begin{table}[tp]
  \centering \small
  \resizebox{\columnwidth}{!}{%
  \begin{tabular}{@{}lcccc@{}}
    \toprule
    Pair & P@1 & CKA & cos-gap & centroid$\downarrow$ \\
    \midrule
    En--Nl & 91/84 & 0.69/0.67 & 0.11/0.09 & 0.36/0.33 \\
    En--Zh & 68/7 & 0.63/0.56 & 0.10/0.03 & 0.48/0.53 \\
    Nl--Zh & 58/6 & 0.60/0.54 & 0.10/0.03 & 0.47/0.53 \\
    \midrule
    MEXA Dutch & \multicolumn{4}{c}{0.87/0.77} \\
    MEXA Chinese & \multicolumn{4}{c}{0.61/0.07} \\
    \bottomrule
  \end{tabular}}
  \caption{Cross-lingual alignment measures beyond retrieval (curriculum ordering).
  Each cell shows the mean CS/non-CS values across seeds at the layer of peak CSLS retrieval in the CS model;
    We measure linear CKA \citep{kornblith2019similarity},
    the cosine gap (mean cosine over the $N$ true translation pairs minus that over a fixed random derangement of the pairs),
    and the per-language centroid distance \citep[lower is better;][]{libovicky2020language}.
  The last two rows are per-language MEXA \citep{kargaran2025mexa} (max over layers vs.\ the English pivot).}
  \label{tab:alignment}
\end{table}

\paragraph{Protocol.}

We embed the FLORES+ \texttt{dev} set \citep{nllb2024scaling, goyal2022flores} (\texttt{eng\_Latn}/\texttt{nld\_Latn}/\texttt{cmn\_Hans}), truncating each sentence to its first $128$ tokens and excluding padding from the mean over positions.
Let $S = (s_{ij})$ be the $N \times N$ matrix of cosine similarities such that $s_{ij}$ is the cosine similarity between the representations of the $i$-th source-language sentence and the $j$-th target-language sentence.
Plain nearest-neighbor retrieval over $S$ suffers from hubness, where a few target sentences are the nearest neighbor of many unrelated queries \citep{radovanovic2010hubs}, so we rank with cross-domain similarity local scaling \citep[$k{=}10$;][]{lample2018word},
forming the matrix $C = (c_{ij})$ where
\[c_{ij}\coloneqq2s_{ij}-r_B(i)-r_A(j),\]
where $r_B(i)$ is the mean cosine between source sentence $i$ and its $k$ nearest targets and $r_A(j)$ the mean cosine between target sentence $j$ and its $k$ nearest sources.
Retrieval is correct for sentence $i$ when its translation ranks first under $C$, and we retrieve in both directions and average:
\[
\mathrm{P@1}\coloneqq\tfrac12\big(\mathrm{acc}_{A\to B}+\mathrm{acc}_{B\to A}\big),
\]
where
\[
\mathrm{acc}_{A\to B}\coloneqq\frac1N\sum_i\mathbf{1}[\arg\max_j c_{ij}=i]
\]
is the fraction of source sentences whose top-ranked target is their own translation, and $\mathrm{acc}_{B\to A}$ is the same in reverse ($C$ is symmetric in the two directions, since $C^{\top}$ is the CSLS matrix of $S^{\top}$).
Chance is $1/N\approx0.1\%$.

\begin{table}[tp]
  \centering \small \setlength{\tabcolsep}{4pt}
  \begin{tabular}{@{}l ccc c@{}}
    \toprule
    & \multicolumn{3}{c}{Bitext retrieval P@1 (\%)} & \\
    \cmidrule(lr){2-4}
    Condition & En--Nl & En--Zh & Nl--Zh & Leaderboard \\
    \midrule
    Shuf & 78.5$\pm$0.7 & 5.4$\pm$0.8 & 4.5$\pm$0.8 & 46.44 \\
    CS/Shuf & 88.8$\pm$0.5 & 65.5$\pm$1.1 & 54.5$\pm$1.0 & 46.55 \\
    Curr & 79.0$\pm$0.8 & 5.5$\pm$0.6 & 4.5$\pm$0.5 & 46.36 \\
    CS/Curr & 87.5$\pm$0.7 & 61.4$\pm$0.8 & 52.5$\pm$0.6 & 46.72 \\
    \midrule
    Word & 85.5$\pm$0.8 & 48.1$\pm$1.5 & 38.5$\pm$1.4 & 46.59 \\
    Sent & 86.1$\pm$0.5 & 46.6$\pm$1.4 & 37.4$\pm$1.5 & 46.70 \\
    Par & 86.0$\pm$0.7 & 25.3$\pm$2.0 & 20.6$\pm$2.0 & 46.56 \\
    Salad & 76.2$\pm$1.2 & 4.2$\pm$0.4 & 3.4$\pm$0.5 & 46.08 \\
    \bottomrule
  \end{tabular}
  \caption{Cross-lingual alignment of all eight conditions at their final checkpoints; alignment columns are $8$-seed mean\,$\pm$\,SD.
  Left: FLORES+ bitext retrieval P@1 (CSLS, $k{=}10$) averaged over layers 6--11, per language pair.
  Right: the leaderboard score of Table~\ref{tab:results}.
  Column abbreviations as in Table~\ref{tab:sweep}.}
  \label{tab:controls-align}\end{table}

\begin{figure*}[tp]
\centering
\includegraphics[width=0.9\textwidth]{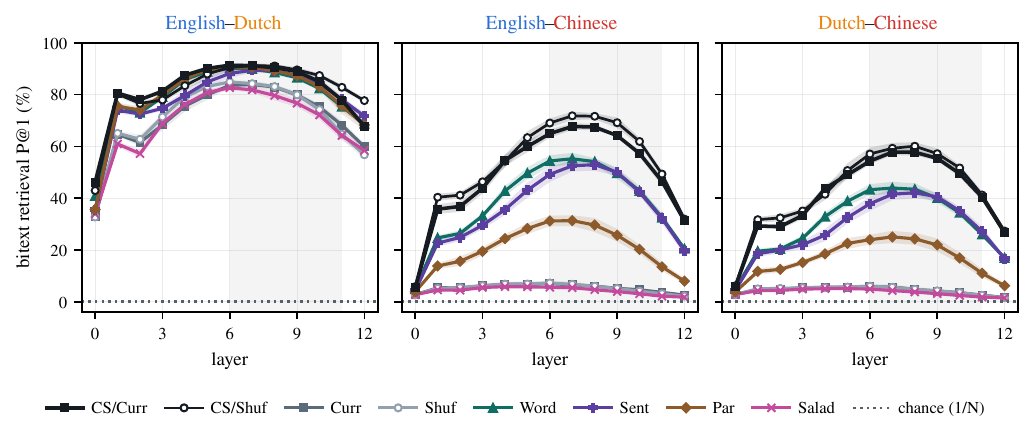}
\caption{Per-layer bitext retrieval P@1 for all eight experimental conditions (\S\ref{sec:setup}), overlaid in each language-pair panel.
We use neutral colors for the four main conditions,
each pairing of corpus (CS/non-CS) and ordering (shuffled/curriculum),
and color for the four shuffled-ordering controls.
Lines are $8$-seed means, bands $\pm1$~SD, dotted line chance.
Shading marks layers $6$--$11$, the range averaged over in Figure~\ref{fig:retrievaltraj}.}
\label{fig:retrieval-all8}
\end{figure*}
\begin{figure*}[tp]
\centering
\includegraphics[width=0.9\textwidth]{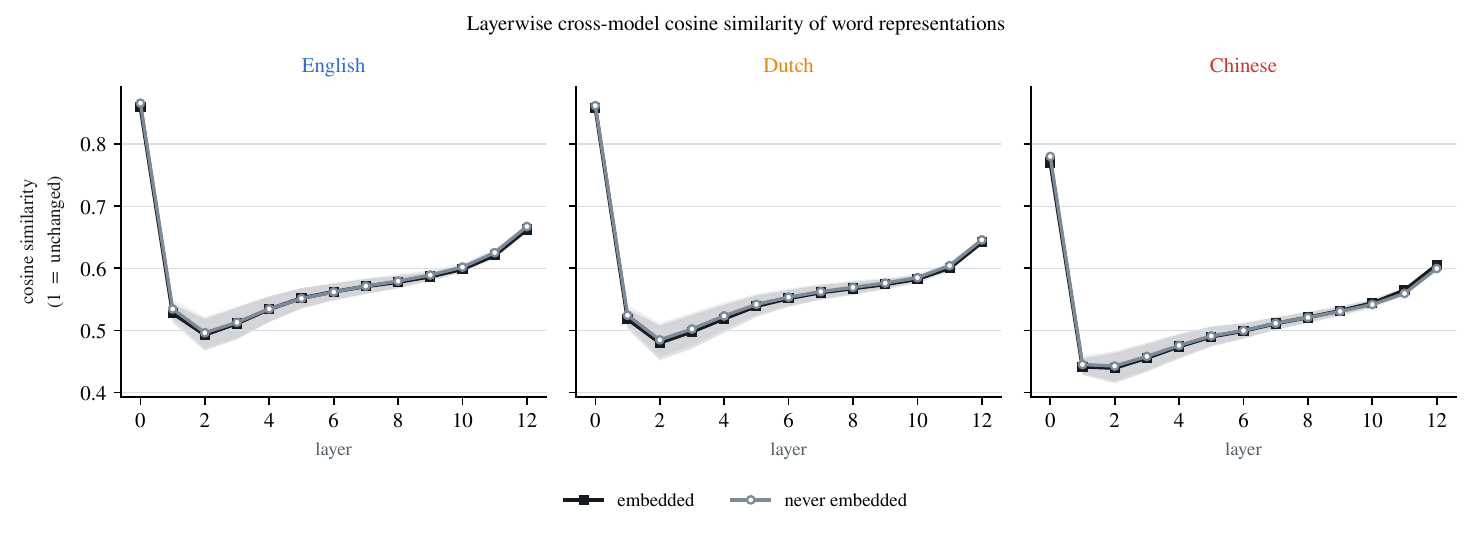}
\caption{Layerwise cosine similarity between the CS and non-CS models' word representations (\S\ref{sec:words}),
shown separately for the embedded words and their never-embedded controls ($8$-seed mean $\pm1$ SD).}
\label{fig:reprdiv}
\end{figure*}

\section{Details on Word-Level Alignment}
\label{app:wordlevel}

\paragraph{Constructing candidate sets for retrieval.}
For each language direction,
we first group words by part of speech and subword-token count.
We bisect each group into embedded and never-embedded words
and match them one-to-one by the Hungarian algorithm on $\log_{10}$ monolingual frequency.
Translation labels come from the MUSE dictionaries \citep{lample2018word}
in addition to the pairs of words substituted during corpus generation.

\paragraph{How code-switching shifts word representations across layers.}
For each of the $2{,}204$ source words in the $1{,}102$ matched pairs of \S\ref{sec:words},
at each layer,
we take the cosine between the CS and non-CS models' representations (Figure~\ref{fig:reprdiv}).
The high cosine similarity between CS and non-CS models at layer zero indicates that code-switching does not affect the word embeddings much. However, early layers seem more affected representationally by the presence of code-switched text in training, as is seen by the low cosine similarity. The representations of words at later layers become again more similar between the CS and non-CS models.

\end{document}